\documentclass{article}
\usepackage{ijcai18}

\usepackage{times}
\usepackage{xcolor}
\usepackage{soul}
\usepackage[utf8]{inputenc}
\usepackage[small]{caption}

\usepackage{graphicx}  
\usepackage{subfigure}
\usepackage{multirow}
\usepackage{booktabs}
\usepackage{diagbox}
\usepackage{color}
\usepackage{bbm}
\usepackage{multicol}
\usepackage{blindtext}
\usepackage{amsmath,amsfonts,amssymb}
\usepackage{algorithm}
\usepackage{algorithmic}
\usepackage{bm}
\usepackage{mathtools}
\usepackage{wrapfig}
\usepackage{url}
\usepackage{amsmath} 

\DeclareMathOperator*{\softmax}{softmax}
\DeclareMathOperator*{\sigmoid}{sigmoid}

\DeclareMathOperator*{\meanpooling}{pooling}
\DeclareMathOperator*{\rss}{RSS}
\DeclareMathOperator*{\rnn}{RNN}

\graphicspath{{./figures/}}

\title{Reinforced Self-Attention Network:\\a Hybrid of Hard and Soft Attention for Sequence Modeling}

\author{
Tao Shen$^1$, 
Tianyi Zhou$^2$, 
Guodong Long$^1$, 
Jing Jiang$^1$, 
Sen Wang$^3$, 
Chengqi Zhang$^1$
\\ 
$^1$ Centre for Artificial Intelligence, School of Software, University of Technology Sydney \\
$^2$ Paul G. Allen School of Computer Science \& Engineering, University of Washington\\
$^3$ School of Information and Communication Technology, Griffith University  \\
tao.shen@student.uts.edu.au, tianyizh@uw.edu, guodong.long@uts.edu.au, \\
jing.jiang@uts.edu.au, sen.wang@griffith.edu.au, chengqi.zhang@uts.edu.au
}
\begin{document}

\maketitle

\begin{abstract}
Many natural language processing tasks solely rely on sparse dependencies between a few tokens in a sentence. Soft attention mechanisms show promising performance in modeling local/global dependencies by soft probabilities between every two tokens, but they are not effective and efficient when applied to long sentences. By contrast, hard attention mechanisms directly select a subset of tokens but are difficult and inefficient to train due to their combinatorial nature. In this paper, we integrate both soft and hard attention into one context fusion model, ``reinforced self-attention (ReSA)'', for the mutual benefit of each other. In ReSA, a hard attention trims a sequence for a soft self-attention to process, while the soft attention feeds reward signals back to facilitate the training of the hard one. For this purpose, we develop a novel hard attention called ``reinforced sequence sampling (RSS)'', selecting tokens in parallel and trained via policy gradient. Using two RSS modules, ReSA efficiently extracts the sparse dependencies between each pair of selected tokens. We finally propose an RNN/CNN-free sentence-encoding model, ``reinforced self-attention network (ReSAN)'', solely based on ReSA.  It achieves state-of-the-art performance on both Stanford Natural Language Inference (SNLI) and Sentences Involving Compositional Knowledge (SICK) datasets. 
\end{abstract}

	\section{Introduction}

Equipping deep neural networks (DNN) with attention mechanisms provides an effective and parallelizable approach for context fusion and sequence compression. It achieves compelling time efficiency and state-of-the-art performance in a broad range of natural language processing (NLP) tasks, such as neural machine translation \cite{bahdanau2015neural,luong2015effective}, dialogue generation \cite{shang2015neural}, machine reading/comprehension \cite{seo2017bidirectional}, natural language inference \cite{liu2016learning}, sentiment classification \cite{li2017adversarial}, etc. Recently, some neural nets based solely on attention, especially self-attention, outperform traditional recurrent \cite{bowman2015snli} or convolutional  \cite{dong2017more} neural networks on NLP tasks, such as machine translation \cite{vaswani2017attention} and sentence embedding \cite{shen2017disan}, which further demonstrates the power of attention mechanisms in capturing contextual dependencies. 

Soft and hard attention are the two main types of attention mechanisms. In soft attention \cite{bahdanau2015neural}, a categorical distribution is calculated over a sequence of elements. The resulting probabilities reflect the importance of each element and are used as weights to produce a context-aware encoding that is the weighted sum of all elements. Hence, soft attention only requires a small number of parameters and less computation time. Moreover, soft attention mechanism is fully differentiable and thus can be easily trained by end-to-end back-propagation when attached to any existing neural net. However, the $\softmax$ function  usually assigns small but non-zero probabilities to trivial elements, which will weaken the attention given to the few truly significant elements. 

Unlike the widely-studied soft attention, in hard attention \cite{xu2015show}, a subset of elements is selected from an input sequence. Hard attention mechanism forces a model to concentrate solely on the important elements, entirely discarding the others. In fact, various NLP tasks solely rely on very sparse tokens from a long text input. Hard attention is well suited to these tasks, because it overcomes the weaknesses associated with soft attention in long sequences. However, hard attention mechanism is time-inefficient with sequential sampling and non-differentiable by virtue of their combinatorial nature. Thus, it cannot be optimized through back-propagation and more typically rely on policy gradient, e.g., REINFORCE \cite{williams1992simple}. As a result, training a hard attention model is usually an inefficient process – some even find convergence difficult – and combining them with other neural nets in an end-to-end manner is problematic.

However, soft and hard attention mechanisms might be integrated into a single model to benefit each other in overcoming their inherent disadvantages, and this notion motivates our study. Specifically, a hard attention mechanism is used to encode rich structural information about the contextual dependencies and trims a long sequence into a much shorter one for a soft attention mechanism to process. Conversely, the soft one is used to provide a stable environment and strong reward signals to help in training the hard one. Such method would improve both the prediction quality of the soft attention mechanism and the trainability of the hard attention mechanism, while boosting the ability to model contextual dependencies. 
To the best of our knowledge, the idea of combining hard and soft attention within a model has not yet been studied.  Existing works focus on only one of the two types.

In this paper, we first propose a novel hard attention mechanism called ``reinforced sequence sampling (RSS)'', which selects tokens from an input sequence in parallel, and differs from existing ones in that it is highly parallelizable without any recurrent structure. We then develop a model,``reinforced self-attention (ReSA)'', which naturally combines the RSS with a soft self-attention. In ReSA, two parameter-untied RSS  are respectively applied to two copies of the input sequence, where the tokens from one and another are called dependent and head tokens, respectively. ReSA only models the sparse dependencies between the head and dependent tokens selected by the two RSS modules. Finally, we build an sentence-encoding model, ``reinforced self-attention network (ReSAN)'',  based on ReSA without any CNN/RNN structure. 

We test ReSAN on natural language inference and semantic relatedness tasks. The results show that ReSAN achieves the best test accuracy among all sentence-encoding models on the official leaderboard of the Stanford Natural Language Inference (SNLI) dataset, and state-of-the-art performance on the Sentences Involving Compositional Knowledge (SICK) dataset. Compared to the commonly-used models, ReSAN is more efficient and has better prediction quality than existing recurrent/convolutional neural networks, self-attention networks, and even well-designed models (e.g., semantic tree or external memory based models). All the experiments codes are released at \url{https://github.com/taoshen58/DiSAN/tree/master/ReSAN}.

\textbf{Notation}: 1) lowercase denotes a vector; 2) bold lowercase denotes a sequence of vectors (stored as a matrix); and 3) uppercase denotes a matrix or a tensor.

\section{Background} \label{sec:background}

\subsection{Attention}
Given an input sequence $\bm{x} = [x_1, \dots, x_n] \in \mathbb{R}^{d_e \times n}$ ($x_i \in \mathbb{R}^{d_e}$ denotes the embedded vector of $i$-th element), and the vector representation of a query $q$, an vanilla \textit{attention} mechanism uses a parameterized compatibility function $f(x_i, q)$ to computes an alignment score between $q$ and each token $x_i$ as the attention of $q$ to $x_i$ \cite{bahdanau2015neural}. A $\softmax$ function is then applied to the alignment scores $a \in \mathbb{R}^{n}$ over all tokens to generate a categorical distribution $p(v|\bm{x}, q)$, where $v=i$ implies that token $x_i$ is selected according to its relevance to query $q$. This can be formally written as 
\begin{align}
&a = \left[f(x_i, q)\right]_{i=1}^n \label{eq:tra_attn_1},\\
&p(v|\bm{x}, q) = \softmax(a) \label{eq:tra_attn_2}.
\end{align}
The output of attention, $s$, is the expectation of sampling a token according to the categorical distribution $p(v|\bm{x}, q)$, i.e.,
\begin{equation}\label{eq:tra_attn_output}
s = \sum_{i=1}^n p(v=i|\bm{x},q)x_i=\mathbb E_{i\sim p(v|\bm{x},q)}[x_i].
\end{equation}

\textit{Multi-dimensional (multi-dim) attention} mechanism \cite{shen2017disan} extends the vanilla one \cite{bahdanau2015neural} to a feature-wise level, i.e., each feature of every token has an alignment score. Hence, rather than a scalar, the output of $f(x_i, q)$ is a vector with the same dimensions as the input, and the resulting alignment scores compose a matrix $a \in \mathbb{R}^{d_e \times n}$. Such feature-level attention has been verified in terms of its ability to capture the subtle variances of different contexts.

\subsection{Self-Attention} \label{sec:self_attn}

Self-attention is a special case of attention where the query $q$ stems from the input sequence itself. Hence, self-attention mechanism can model the dependencies between tokens from the same sequence. Recently, a variety of self-attention mechanisms have been developed, each serving a distinct purpose, but most can be roughly categorized into two types, token2token self-attention and source2token self-attention. 

\textbf{Token2token self-attention} mechanisms aim to produce a context-aware representation for each token in light of its dependencies on other tokens in the same sequence. The query $q$ is replaced with the token $x_j$, and the dependency of $x_j$ on another token $x_i$ is computed by $f(x_i, x_j)$. There are two proposed self-attentions in this type, i.e., scaled dot-product attention which composes the multi-head attention \cite{vaswani2017attention} and masked self-attention which leads to directional self-attention \cite{shen2017disan}. Because the latter experimentally outperforms the former, we select the masked self-attention as our fundamental soft self-attention module.

\textit{Masked Self-Attention} is more sophisticated than scaled dot-product attention in that, it uses multi-dim and multi-layer perceptron with an additional position mask, rather than a scaled dot-product, as the compatibility function, i.e.,
\begin{align}\label{eq:masked_self_attn}
\notag &f(x_i, x_j)=\\ 
&c\cdot\tanh\left([W^{(1)}x_i+W^{(2)}x_j+b^{(1)}]/c\right) +M_{ij},
\end{align}
where $c$ is a scalar and $M$ is the mask with each entry $M_{ij}\in\{-\infty, 0\}$. When $M_{ij}=-\infty$, applying the $\softmax$ function to $a$ results in a zero probability, $p(z=i|\bm{x}, x_j)=0$, which switches off the attention of $x_j$ to $x_i$. An asymmetric mask where $M_{ij}\neq M_{ji}$ enforces directional attention between $x_i$ and $x_j$, which can encode temporal order information. Two positional masks have been designed to encode the forward and backward temporal order, respectively, i.e.,
\begin{equation}\label{eq:fw_bw_masks} 
\notag
M_{ij}^{fw} = \left\{
\begin{array}{ll}
0,& i < j\\
- \infty,& \text{otherwise}\\
\end{array}
\right.
\hspace{1em}
M_{ij}^{bw} =\left\{\begin{array}{ll}
0, & i > j \\
- \infty, & \text{otherwise}\\
\end{array}\right.
\end{equation}
In forward and backward masks, $M_{ii}=-\infty$. Thus, the attention of a token to itself is blocked, so the output of masked self-attention mechanism comprises the features of the context around each token rather than context-aware features. 

\textit{Directional self-attention} uses a fusion gate to combine the embedding of each token with its context. Specifically, a fusion gate combines the input and output of a masked self-attention to produce context-aware representations. This idea is similar to the highway network \cite{srivastava2015highway}. 

\textbf{Source2token self-attention} mechanisms \cite{shen2017disan} remove the query $q$ from the compatibility function in Eq.(\ref{eq:tra_attn_1}) and directly compresses a sequence into a vector representation calculated from the dependency between each token $x_i$ and the entire input sequence $\bm{x}$. Hence, this form of self-attention is highly data- and task- driven. 

\section{Proposed Models}

This section begins by introducing a hard attention mechanism called RSS in Section \ref{sec:hard_attention}, followed by integrating the RSS with a soft self-attention mechanism into a context fusion model called ReSA in Section \ref{sec:resa}. Finally, a model named ReSAN, based on ReSA, is designed for sentence encoding tasks in Section \ref{sec:applications}

\subsection{Reinforced Sequence Sampling (RSS)} \label{sec:hard_attention}

The goal of hard attention mechanism is to select a subset of critical tokens that provides sufficient information to complete downstream tasks, so any further computations on the trivial tokens can be saved. In the following, we introduce a hard attention mechanism called RSS. Given an input sequence $\bm{x} = [x_1, \dots, x_n]$, RSS generates an equal-length sequence of binary random variables $\bm{z} = [z_1, \dots, z_n]$ where $z_i =1$ implies that $x_i$ is selected whereas $z_i =0$ indicates that $x_i$ is discarded. In RSS, the elements of $\bm{z}$ are sampled in parallel according to probabilities computed by a learned attention mechanism. This is more efficient than using MCMC with iterative sampling. The particular aim of RSS is to learn the following product distribution.
\begin{align} \label{eq:z_seq_prob}
&p(\bm{z}|\bm{x}; \theta_r) = \prod_{i=1}^{n} p(z_i|\bm{x}; \theta_r), \\ 
\notag \text{where } & p(z_i|\bm{x}; \theta_r) = g(f(\bm{x}; \theta_f)_i; \theta_g).
\end{align}
The function $f(\cdot; \theta_f)$ denotes a context fusion layer, e.g., Bi-LSTM, Bi-GRU, etc., producing context-aware representation for each $x_i$. Then, $g(\cdot; \theta_g)$ maps $f(\cdot; \theta_f)$ to the probability of selecting the token. Note we can sample all $z_i$ for different $i$ in parallel because the probability of $z_i$ (i.e., whether $x_i$ is selected) does not depends on $z_{i-1}$. This is because the context features given by $f(\cdot; \theta_f)$ already take the sequential information into account, so the conditionally independent sampling does not discard any useful information. 

To fully explore the high parallelizability of attention, we avoid using recurrent models in this paper. Instead we apply a more efficient $f(\cdot; \theta_f)$ inspired by source2token self-attention and intra-attention \cite{liu2016learning}, i.e., 
\begin{align} 
& f(\bm{x}; \theta_f)_i = [x_i; \meanpooling(\bm{x}); x_i \odot \meanpooling(\bm{x})], \label{eq:function_f}\\ 
& g(h_i; \theta_g) =  \sigmoid(w^T\sigma(W^{(R)}h_i + b^{(R)}) + b), \label{eq:function_g}
\end{align}
where $\odot$ denotes the element-wise product, and the $\meanpooling(\cdot)$ represents the mean-pooling operation along the sequential axis. RSS selects a subset of tokens by sampling $z_i$ according to the probability given by $g(h_i; \theta_g)$ for all $i=1,2,\dots,n$ in parallel.

For the training of RSS, there are no ground truth labels to indicate whether or not a token should be selected, and the discrete random variables in $\bm{z}$ lead to a non-differentiable objective function. Therefore, we formulate learning the RSS parameter $\theta_r$ as a reinforcement learning problem, and apply the policy gradient method. Further details on the model training are presented in Section \ref{sec:model_training}. 

\subsection{Reinforced Self-Attention (ReSA)} \label{sec:resa}

The fundamental idea behind this paper is that the hard and soft attention mechanisms can mutually benefit each other to overcome their inherent disadvantages via interaction within an integrated model. Based on this idea, we develop a novel self-attention termed ReSA. On the one hand, the proposed RSS provides a sparse mask to a self-attention module that only needs to model the dependencies for the selected token pairs. Hence, heavy memory loads and computations associated with soft self-attention can be effectively relieved. On the other hand, ReSA uses the output of the soft self-attention module for prediction, whose correctness (as compared to the ground truth) is used as reward signal to train the RSS. This alleviates the difficulty of training hard attention module. 

\begin{figure}[t] 
	\centering
	\includegraphics[width=0.4\textwidth]{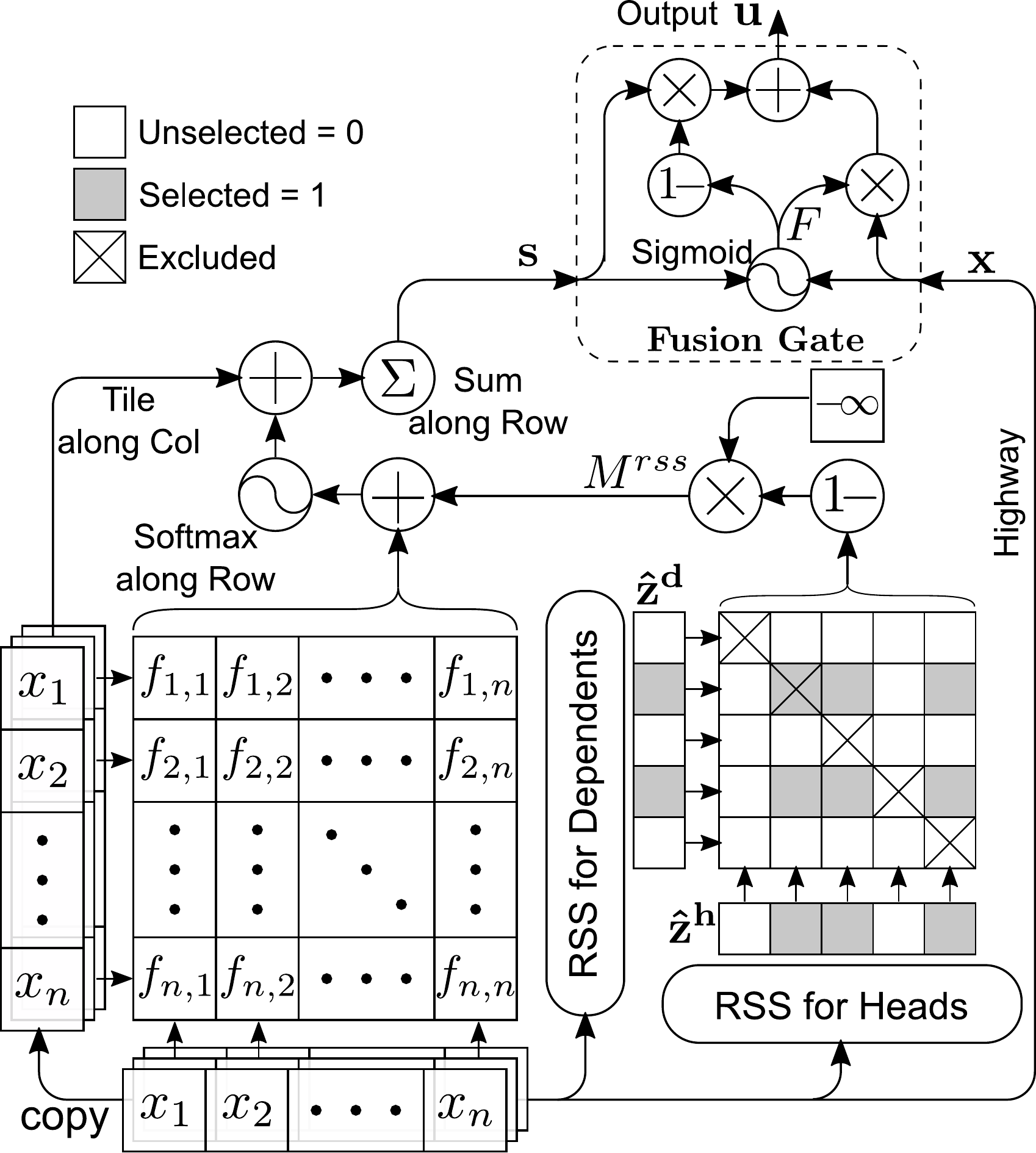}
	\caption{\small Reinforced self-attention (ReSA) model. $f_{i,j}$ denotes the alignment score obtained from $f(x_i, x_j)$.}
	\label{fig:resa} 
	\centering
\end{figure}

Figure \ref{fig:resa} shows the detailed architecture of ReSA. Given the token embedding in an input sequence, $\bm{x} = [x_1,  \dots, x_n]$, ReSA aims to produce token-wise context-aware representations, $\bm{u} = [u_1,  \dots, u_n]$. Unlike previous self-attention mechanisms, ReSA only selects a subset of head tokens, and generates their context-aware representations by only relating each head token to a small subset of dependent tokens. This notion is based on the observation that for many NLP tasks, the final prediction only relies on a small set of key words and their contexts, and each key word only depends on a small set of other words. Namely, the dependencies between tokens from the same sequence are sparse. 

In ReSA, we use two RSS modules, as outlined in Section \ref{sec:hard_attention}, to generate two sequences of labels for the selections of head and dependent tokens, respectively, i.e., 
\begin{align} 
& \bm{\hat{z}^h} = [\hat{z}^h_1, \dots, \hat{z}^h_n] \sim \rss(\bm{x}; \theta_{rh}), \label{eq:ssr_head}\\ 
& \bm{\hat{z}^d} = [\hat{z}^d_1, \dots, \hat{z}^d_n] \sim \rss(\bm{x}; \theta_{rd}), \label{eq:ssr_dep}
\end{align}
We use $\bm{\hat{z}^h}$ and $\bm{\hat{z}^d}$ sampled from the two independent (parameter untied) RSS to generate an $n\times n$ mask $M^{rss}$, i.e., 
\begin{equation}\label{eq:mask_rss} 
M_{ij}^{rss} = \left\{\begin{matrix}
0,& \hat{z}^d_i = \hat{z}^h_j = 1 \text{ \& } i \neq j \\  
- \infty,& \text{otherwise}.
\end{matrix}\right. 
\end{equation}
The resulting mask is then applied as an extra mask to the masked self-attention mechanism introduced in Section \ref{sec:self_attn}. Specifically, we add $M^{rss}$ to Eq.(\ref{eq:masked_self_attn}) and use
\begin{equation}\label{eq:comp_func_rss} 
f^{rss}(x_i, x_j) = f(x_i, x_j) + M^{rss}_{ij}
\end{equation}
to generate the alignment scores. For each head token $x_j$, a $\softmax$ function is applied to $f^{rss}(\cdot, x_j)$, which produces a categorical distribution over all dependent tokens, i.e.,
\begin{equation}
P^j = \softmax([f^{rss}(x_i, x_j)]_{i=1}^{n}), \text{ for } j= 1,\dots, n. 
\end{equation}
The context features of $x_j$ is computed by
\begin{equation}
s_j = \sum_{i=1}^{n} P^j_i \odot x_i, \text{ for } j = 1,\dots, n,
\end{equation}
where $\odot $ denotes a broadcast product in the vanilla attention or an element-wise product in the multi-dim attention. 

For a \textit{selected} head token, as formulated in Eq.(\ref{eq:mask_rss}), the attention from a token to itself is disabled in $M^{rss}$, so the $s_j$ for the selected head token encodes only the context features but not the desired context-ware embedding. For an \textit{unselected} head token $x_j$ with $\hat{z}^h_j=0$, its alignment scores over all dependent tokens are equal to $-\infty$, which leads to the equal probabilities in $P^j$ produced by the $\softmax$ function. Hence, $s_j$ for each unselected token $x_j$ can be regarded as the result of mean-pooling over all dependent tokens. 

To merge the word embedding with its context feature for the \textit{selected} heads, and distinguish the representations from others for the \textit{unselected} heads, a  fusion gate is used to combine $\bm{s}$ with the input embedding $\bm{x}$ in parallel and generate the final context-aware representations for all tokens, i.e., 
\begin{align}
&F = \sigmoid \left(W^{(f)}[\bm{x}; \bm{s}] + b^{(f)}\right), \label{eq:context_fusion1} \\
&\bm{u} =F \odot \bm{x} +  (1 - F)  \odot \bm{s}, \label{eq:context_fusion2}
\end{align}
where $W^{(f)}$, $b^{(f)}$ are the learnable parameters. The context-aware representations, $\bm{u} = [u_1, \dots, u_n]$, are final output. 
One primary advantage of ReSA is that it generates better predictions using less time and memory than existing self-attention mechanisms. In particular, major computations of ReSA are 1) the inference of self-attention over a shorter subsequence, and 2) the mean-pooling over the remaining elements. This is much more time- and memory- efficient than computing the self-attention over the entire input sequence. 

\subsection{Applications of the Proposed Models} \label{sec:applications}

To adapt ReSA for sentence encoding tasks, we build an RNN/CNN-free network, called reinforced self-attention network (ReSAN), which is solely based on ReSA and source2token self-attention (Section \ref{sec:self_attn}). In particular, we pass the output sequence of ReSA into a source2token self-attention module to generate a compressed vector representation, $e \in \mathbb{R}^{d_e}$, which encodes the semantic and syntactic knowledge of the input sentence and can be used for various downstream NLP tasks.

Further, we propose two simplified variants of ReSAN with a simpler structure or fewer parameters, i.e., 1) \textbf{ReSAN w/o unselected heads} which only applies the soft self-attention to the selected head and dependent tokens, and 2) \textbf{ReSAN w/o dependency restricted} which use only one RSS to select tokens for both heads and dependents. Both variants entirely discard the information of the unselected tokens and hence are more time-efficient. However, neither can be used for context fusion, because the input and output sequences are not equal in length.

\section{Model Training} \label{sec:model_training}

The parameters in ReSAN can be divided into two parts, $\theta_r$ for the RSS modules and $\theta_s$ for the rest parts which includes word embeddings, soft self-attention module, and classification/regression layers. Learning $\theta_s$ is straightforward and can be completed by back-propagation in an end-to-end manner. However, Optimizing $\theta_r$ is more challenging because the RSS modules contain discrete variables $\bm{z}$ and, thus, the objective function is non-differentiable w.r.t. $\theta_r$. 

In supervised classification settings, we use the cross-entropy loss plus L2 regularization penalty as the loss, i.e., 
\begin{equation}
J_s(\theta_s) = \mathbb{E}_{(\bm{x^*}, y^*)\sim \mathcal{D}}[- \log p(y=y^*|\bm{x^*}; \theta_{s,r})]  + \gamma \| \theta_s\|^2,
\end{equation}
where $(\bm{x^*}, y^*)$ denotes a sample from dataset $\mathcal{D}$. The loss above is used for learning $\theta_s$ by back-propagation algorithm.

\begin{table*}[htbp] \small
	\centering
	\setlength{\tabcolsep}{2.5pt}
	\begin{tabular}{@{}lccccc@{}} 
		\toprule
		\textbf{Model}& \textbf{$\boldsymbol{|\theta|}$} & \textbf{T(s)/epoch}  & \textbf{Inference T(s)}&\textbf{Train Accuracy} & \textbf{Test Accuracy} \\ \midrule
		300D LSTM encoders \cite{bowman2016fast}&            3.0m&                &  &               83.9&               80.6\\ 
		300D SPINN-PI encoders \cite{bowman2016fast}&            3.7m&              &  &                   89.2&               83.2\\ 
		600D Bi-LSTM encoders \cite{liu2016learning}&            2.0m&                 &  &               86.4&               83.3\\ 
		600D Bi-LSTM +intra-attention \cite{liu2016learning}&            2.8m&               &  &                 84.5&               84.2\\ 
		300D NSE encoders  \cite{munkhdalai2016neural_2}&            3.0m&                 &  &                86.2&               84.6\\
		600D Deep Gated Attn. \cite{chen2017recurrent}&           11.6m&              &  &                   90.5&               85.5\\
		600D Gumbel TreeLSTM encoders \cite{choi2017Learning} &           10m&                 &  &                93.1&               86.0\\
		600D Residual stacked encoders \cite{nie2017shortcut} &           29m&           &  &                     91.0&               86.0\\\midrule
		Bi-LSTM \cite{graves2013hybrid}&            2.9m&        2080&	9.2&                     90.4&             85.0\\ 
		Bi-GRU \cite{chung2014empirical}&            2.5m&          1728&	9.3&            91.9&             84.9\\ 
		Multi-window CNN \cite{kim2014convolutional}&            1.4m&       284 &	2.4&                 89.3&              83.2\\ 
		Hierarchical CNN \cite{gehring2017convolutional} &            3.4m&           343&	2.9&             91.3&              83.9\\ 
		Multi-head \cite{vaswani2017attention}&            2.0m&             345&	3.0&                  89.6&               84.2\\ 
		DiSAN \cite{shen2017disan}&            2.4m&             587&	7.0&             91.1&               85.6\\ \midrule
		300D ReSAN&            3.1m&            622&	5.5&              92.6&               \textbf{86.3}\\ \bottomrule
	\end{tabular}
	\caption{Experimental results for different methods on SNLI. \textbf{$\boldsymbol{|\theta|}$}: the number of parameters (excluding word embedding part). \textbf{T(s)/epoch}: average training time (second) per epoch. \textbf{Inference T(s)}: average inference time (second) for all dev data on SNLI with a batch size of $100$.} 
	\label{table:exps_snli_accu}
\end{table*}

Optimizing $\theta_r$ is formulated as a reinforcement learning problem solved by the policy gradient method (i.e., REINFORCE algorithm). In particular, RSS plays as an agent and takes \textit{action} of whether to select a token or not. After going through the entire sequence, it receives a loss value from the classification problem, which can be regarded as the negative delay \textit{reward} to train the agent. Since the overall goal of RSS is to select a small subset of tokens for better efficiency and meanwhile retain useful information, a penalty limiting the number of selected tokens is included in the reward $\mathcal{R}$, i.e., 
\begin{equation}
\mathcal{R} = \log p(y=y^*|\bm{x^*}; \theta_s, \theta_r) - \lambda \sum \hat{z}_i / len(\bm{x^*}),
\end{equation}
where $\lambda$ is the penalty weight and is fine-tuned with values from $\{0.005, 0.01, 0.02\}$ in all experiments. Then, the objective of learning $\theta _r$ is to maximize the expected reward, i.e.,
\begin{equation} 
J_r(\theta_r)=\mathbb{E}_{(\bm{x^*}, y^*)\sim \mathcal{D}}\{\mathbb{E}_{\bm{\hat{z}}}[\mathcal{R}]\}
\approx \dfrac{1}{N} \sum_{\bm{x^*}, y^*}  \mathbb{E}_{\bm{\hat{z}}}[\mathcal{R}]
\end{equation}
where the $\bm{\hat{z}} = (\bm{\hat{z}^h},  \bm{\hat{z}^d}) \sim p(\bm{z^h}|\bm{x^*}; \theta_{rh})  p(\bm{z^d}|\bm{x^*}; \theta_{rd}) \triangleq \pi (\bm{\hat{z}}; \bm{x^*}; \theta_r)$ and $N$ is sample number in the dataset. Based on  REINFORCE, the policy gradient of $J_r(\theta_r)$ w.r.t $\theta_r$ is
\begin{align} 
&\bigtriangledown_{\theta_r} J_r(\theta_r) = \dfrac{1}{N} \sum_{\bm{x^*}, y^*} \sum_{\bm{\hat{z}}} \mathcal{R} \bigtriangledown_{\theta_r} \pi (\bm{\hat{z}}; \bm{x^*}; \theta_r) \\
& = \dfrac{1}{N} \sum_{\bm{x^*}, y^*}  \mathbb{E}_{\bm{\hat{z}}}[\mathcal{R} \bigtriangledown_{\theta_r} \log \pi (\bm{\hat{z}}; \bm{x^*}; \theta_r)].
\end{align}

Although theoretically feasible, it is not practical to optimize $\theta_s$ and $\theta_r$ simultaneously, since the neural nets cannot provide accurate reward feedback to the hard attention at the beginning of the training phrase. Therefore, in early stage, the RSS modules are not updated, but rather forced to select all tokens (i.e., $\bm{z}=\bm{1}$ ). And, $\theta_s$ is optimized for several beginning epochs until the loss over development set does not decrease significantly. The resulting ReSAN now can provide a solid environment for training RSS modules through reinforcement learning. $\theta_r$ and $\theta_s$ are then optimized simultaneously to pursue a better performance by selecting critical token pairs and exploring their dependencies.

\textbf{Training Setup:} All experiments are conducted in Python with Tensorflow and run on a Nvidia GTX 1080Ti. We use Adadelta as optimizer, which performs more stable than Adam on ReSAN. All weight matrices are initialized by Glorot Initialization \cite{glorot2010understanding} and the biases are initialized as zeros. We use 300D GloVe 6B pre-trained vectors \cite{pennington2014glove} to initialize the word embeddings \cite{liu2018semantic}. The words which do not appear in GloVe from the training set are initialized by sampling from uniform distribution between $[-0.05,0.05]$. We choose Dropout \cite{srivastava2014dropout} keep probability from $\{0.65, 0.70, 0.75, 0.8\}$ for all models and report the best result. The weight decay factor $\gamma$ for L2 regularization is set to $5\times 10^{-5}$. The number of hidden units is $300$. 

\section{Experiments}

We implement ReSAN, its variants and baselines on two NLP tasks, language inference in Section \ref{sec:nli} and semantic relatedness in Section \ref{sec:semantic_relatedness}. A case study is then given to provide the insights into model. 

The baselines are listed as follows: 
\textbf{1) Bi-LSTM}: 600D bi-directional LSTM (300D forward LSTM + 300D backward LSTM) \cite{graves2013hybrid}; 
\textbf{2) Bi-GRU}: 600D bi-directional GRU \cite{chung2014empirical}; 
\textbf{3) Multi-window CNN}:  600D CNN sentence embedding model (200D for each of 3, 4, 5-gram) \cite{kim2014convolutional}; 
\textbf{4) Hierarchical CNN}: 3-layer 300D CNN \cite{gehring2017convolutional} with kernel length 5. GLU \cite{dauphin2016language} and residual connection \cite{he2016deep} are applied;
\textbf{5) Multi-head}: 600D multi-head attention (8 heads, each has 75 hidden units), where the positional encoding method is applied to the input \cite{vaswani2017attention};
\textbf{6) DiSAN}: 600D directional self-attention network (forward+backward masked self-attn) \cite{shen2017disan}.

\subsection{Natural Language Inference} \label{sec:nli}

The goal of natural language inference is to infer the semantic relationship between a pair of sentences, i.e., a premise and the corresponding hypothesis. The possible relationships are \textit{entailment}, \textit{neutral} or \textit{contradiction}. This experiment is conducted on the Stanford Natural Language Inference \cite{bowman2015snli} (SNLI) dataset which consists of 549,367/9,842/9,824 samples for training/dev/test.

In order to apply sentence encoding model to SNLI, we follow  \citeauthor{bowman2016fast}~\shortcite{bowman2016fast} and use two parameter-tied sentence encoding models to respectively produce the premise and the hypothesis encodings, i.e., $s^p$, $s^h$. Their semantic relationship is represented by the concatenation of $s^p$, $s^h$, $s^p\!-\!s^h$ and $s^p\!\odot\! s^h$, which is passed to a classification module to generate a categorical distribution over the three classes. 

The experimental results for different methods from leaderboard and our baselines are shown in Table \ref{table:exps_snli_accu}. Compared to the methods from official leaderboard, ReSAN outperforms all the sentence encoding based methods and achieves the best test accuracy. Specifically, compared to the last best models, i.e., 600D Gumbel TreeLSTM encoders and 600D Residual stacked encoders, ReSAN uses far fewer parameters with better performance. Moreover, in contrast to the RNN/CNN based models with attention or memory module, ReSAN uses attention-only modules with equal or fewer parameters but outperforms them by a large margin, e.g., 600D Bi-LSTM + intra-attention (+3.0\%), 300D NSE encoders (+1.7\%) and 600D Deep Gated Attn (+0.8\%). Furthermore, ReSAN even outperforms the 300D SPINN-PI encoders by 3.1\%., which is a recursive model and uses the result of an external semantic parsing tree as an extra input.

In addition, we compare ReSAN with recurrent, convolutional, and attention-only baseline models in terms of the number of parameters, training/inference time and test accuracy. Compared to the recurrent models (e.g., Bi-LSTM and Bi-GRU), ReSAN shows better prediction quality and more compelling efficiency due to parallelizable computations. Compared to the convolutional models (i.e., Multi-window CNN and Hierarchical CNN), ReSAN significantly outperforms them by 3.1\% and 2.4\% respectively due to the weakness of CNNs in modeling long-range dependencies. Compared to the attention-based models,  multi-head attention and DiSAN, ReSAN uses a similar number of  parameters with better test performance and less time cost.

\begin{table}[t] \small 
	\centering
	\setlength{\tabcolsep}{1pt}
	\begin{tabular}{@{}lccc@{}}
		\toprule
		\textbf{Model}& \textbf{$\boldsymbol{|\theta|}$} & \textbf{Inference T(s)}& \textbf{Test Accu.}\\ \midrule
		ReSAN  &            3.1m&       5.5&        86.3\\
		ReSAN w/o unselected heads&            3.1m&        5.3&       86.1\\
		ReSAN w/o dependency restricted&           2.8m&          4.6&     85.6\\
		ReSAN w/o hard attention  &            2.5m&       7.0&        86.0\\
		ReSAN w/o soft self-attention &            1.0m&       1.6&        83.4\\
		ReSAN w/o all attentions & 0.5m&     1.8&       83.1\\ \bottomrule
	\end{tabular}
	\caption{\small An ablation study of ReSAN.}
	\label{table:exps_snli_ablation}	
\end{table}

Further, we conduct an ablation study of ReSAN, as shown in Table \ref{table:exps_snli_ablation}, to evaluate the contribution of each component. One by one, each component is removed and the changes in test accuracy are recorded.  In addition to the two variants of ReSAN introduced in Section \ref{sec:applications}, we also remove 1) the hard attention module, 2) soft self-attention module and 3) both hard attention and soft self-attention modules. In terms of prediction quality, the results show that 1) the unselected head tokens do contribute to the prediction, bringing 0.2\% improvement; 2) using separate RSS modules to select the head and dependent tokens improves accuracy by 0.5\%; and 3) hard attention and soft self-attention modules improve the accuracy by 0.3\% and 2.9\% respectively. In terms of inference time, it shows that 1) the two variants are more time-efficient but have poorer performance; and 2) applying the RSS modules to self-attention or attention improves not only performance but also time efficiency.

\subsection{Semantic Relatedness} \label{sec:semantic_relatedness}

Semantic relatedness aims to predict the similarity degree of a given pair of sentences, which is formulated as a regression problem. We use $s^1$ and $s^2$ to denote the encodings of the two sentences, and assume the similarity degree is a scalar between $[1, K]$. Following  \citeauthor{tai2015improved}~\shortcite{tai2015improved}, the relationship between the two sentences is represented as a concatenation of $s^1 \! \odot \! s^2$ and $|s^1\! -\! s^2|$. The representation is fed into a classification module with $K$-way categorical distribution output. We implement ReSAN and baselines on the Sentences Involving Compositional Knowledge \cite{marelli2014sick} (SICK) dataset, which provides the ground truth as similarity degree between $[1, 5]$. SICK come with a standard training/dev/test split of 4,500/500/4,927 samples. 

\begin{table}[t] \small 
	\centering
	\setlength{\tabcolsep}{1pt}
	\begin{tabular}{@{}lccc@{}}
		\toprule
		\textbf{Model}& \textbf{Pearson's $\bm{r}$} &\textbf{Spearman's $\bm{\rho}$}& \textbf{MSE}  \\ \midrule
		Meaning Factory$^a$&            .8268&               .7721&                .3224\\
		ECNU$^b$&            .8414&               /&                /\\
		DT-RNN$^c$&            .7923 (.0070)&               .7319 (.0071)&               .3822 (.0137) \\ 
		SDT-RNN$^c$&            .7900 (.0042)&               .7304 (.0042)&                .3848 (.0042)\\ 
		Cons. Tree-LSTM$^d$&            .8582 (.0038)&               .7966 (.0053)&                .2734 (.0108)\\ 
		Dep. Tree-LSTM$^d$&            .8676 (.0030)&               .8083 (.0042)&                \textbf{.2532 (.0052)}\\ \midrule
		Bi-LSTM& .8473 (.0013)& .7913 (.0019)& .3276 (.0087)\\
		Bi-GRU& .8572 (.0022)& .8026 (.0014)& .3079 (.0069)\\
		Multi-window CNN& .8374 (.0021)& .7793 (.0028)& .3395 (.0086)\\
		Hierarchical CNN& .8436 (.0014)& .7874 (.0022)& .3162 (.0058)\\
		Multi-head& .8521 (.0013)& .7942 (.0050)& .3258 (.0149)\\
		DiSAN& .8695 (.0012)&  .8139 (.0012)& .2879 (.0036)\\ \midrule
		ReSAN& \textbf{.8720 (.0014)}& \textbf{.8163 (.0018)}& \textit{.2623 (.0053)}\\ \bottomrule
	\end{tabular}
	\caption{ \small Experimental results for different methods on SICK semantic relatedness dataset. The reported accuracies are the mean of five runs (standard deviations in parentheses). Cons. and Dep. represent Constituency and Dependency, respectively. $^a$\protect\cite{bjerva2014meaning}, $^b$\protect\cite{zhao2014ecnu}, $^c$\protect\cite{socher2014grounded}, $^d$\protect\cite{tai2015improved}}
	\label{table:exps_sick_accu}
\end{table}

The results in Table \ref{table:exps_sick_accu} show that the ReSAN achieves state-of-the-art or competitive performance for all three metrics. Particularly, ReSAN outperforms the feature engineering method by a large margin, e.g., Meaning Factory and ECNU. ReSAN also significantly outperforms the recursive models, which is widely used in semantic relatedness task, especially ones that demand external parsing results, e.g., DT/SDT-RNN and Tree-LSTMs. Further, ReSAN achieves the best results among all the recurrent, convolutional and self-attention models listed as baselines. This thoroughly demonstrates the capability of ReSAN in context fusion and sentence encoding.

\subsection{Case Study}
To gain an insights into how the hard/soft attention and fusion gate work within ReSA, we visualize their resulting values in this section. Note that only the values at token level are illustrated. If the attention probabilities and the gate values are feature-level, we average the probabilities over all features. 

Two sentences from the SNLI test set serve as examples for this case study: 1) ``\textit{The three men sit and talk about their lives.}'' and 2) ``\textit{A group of adults are waiting for an event.}''.

\begin{figure}[t] 
	\centering
	\subfigure[Sentence 1]{
		\label{fig:case_soft_attn_s1}
		\includegraphics[width=0.22\textwidth]{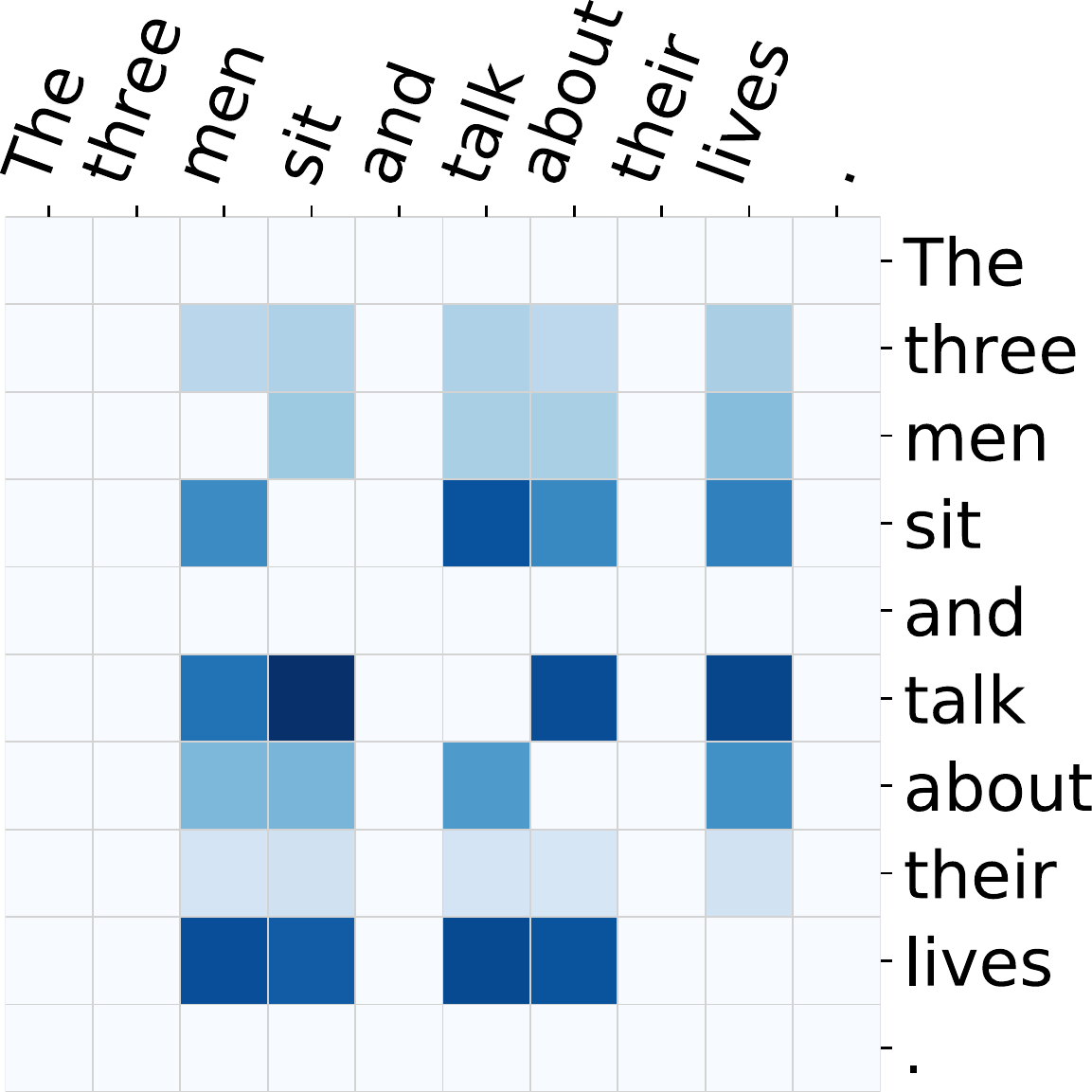}}
	\subfigure[Sentence 2]{
		\label{fig:case_soft_attn_s2} 
		\includegraphics[width=0.22\textwidth]{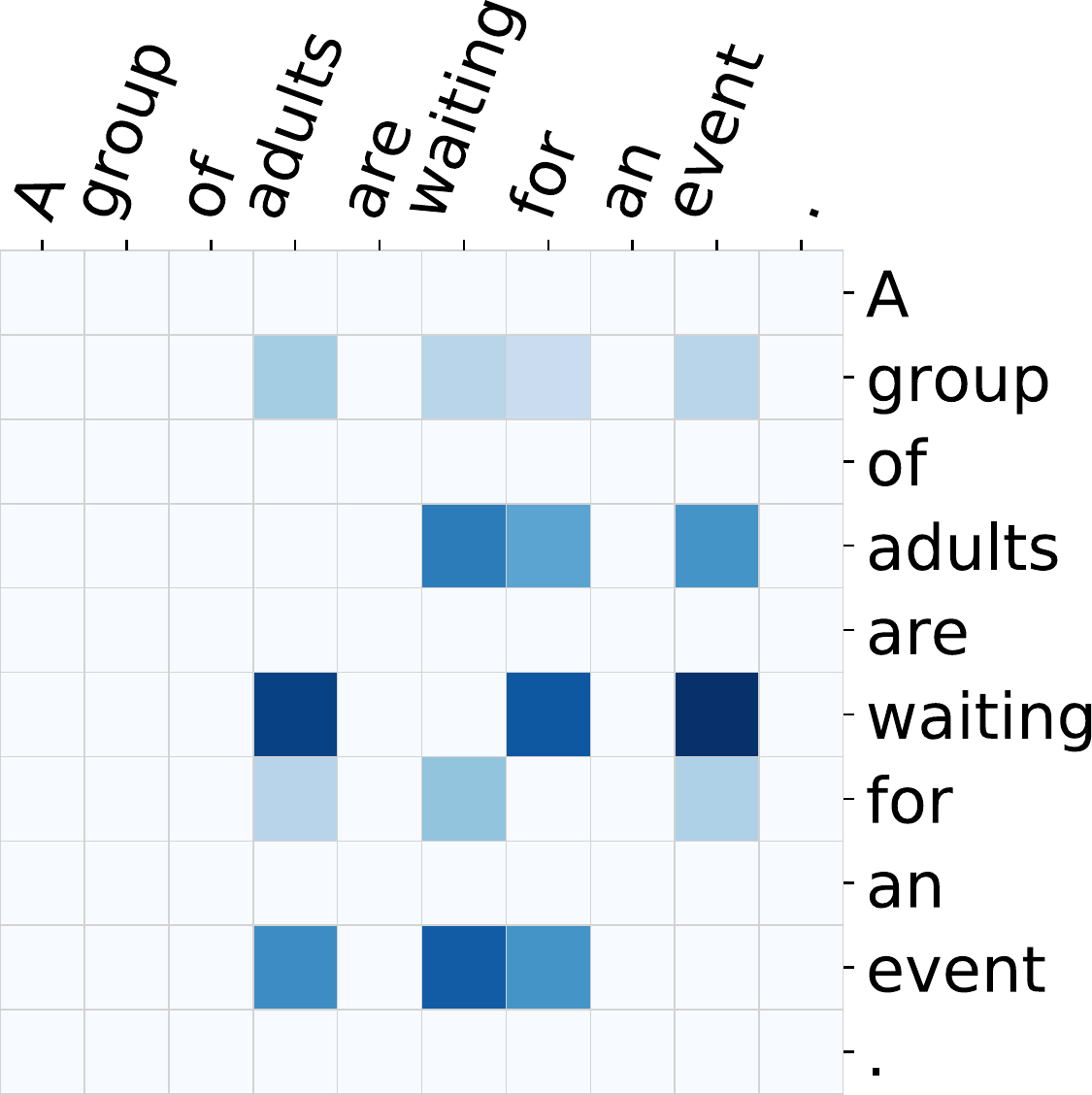}}
	\caption{Attention probabilities of soft self-attention in ReSA. The tokens aligned in horizontal axis are heads, and the tokens aligned in vertical axis are dependents.}
	\label{fig:case_soft_attn}
\end{figure}

The head and dependent tokens selected by RSS modules are show in Figure \ref{fig:case_soft_attn} (a small square with color white denotes unselection and vice versa). It shows that more dependent tokens are selected than the head tokens, because all non-trivial dependents should be retained to adequately modify the corresponding heads, e.g., \textit{three, their} in sentence 1 and \textit{group} in sentence 2, whereas only the key heads should be kept to compose the trunk of a sentence. It also shows that most stop words (i.e., articles, conjunctions, prepositions, etc.) are selected as neither head tokens nor dependent tokens.

We also visualize the probability distributions of the soft self-attention module in Figure \ref{fig:case_soft_attn} (the depth of color blue). From the figure, we observe that 1) the semantically important words (e.g., noun and verb) usually receive great attention from all the other tokens, e.g., \textit{sit, talk, lives} in sentence 1 and \textit{adults, waiting, event} in sentence 2; and 2) the attention score increases if the token pair can be constituted to a sense-group, e.g., (\textit{sit}, \textit{talk}) in sentence 1 and (\textit{adults}, \textit{waiting}), (\textit{waiting}, \textit{event}) in sentence 2.

\section{Related Work} \label{ext:related}

Applying reinforcement learning (RL) to natural language processing (NLP) tasks recently attracts enormous interests for two main purposes, i.e., optimizing the model according to non-differentiable objectives and accelerating the model speed. 
\citeauthor{lei2016rationalizing}~\shortcite{lei2016rationalizing} propose a method to select a subset of a review passage for sentiment analysis from a specific aspect.
\citeauthor{he2016dual}~\shortcite{he2016dual} use RL method to fine-tune a bilingual machine translation model by well-trained monolingual language models.
\citeauthor{yogatama2016learning}~\shortcite{yogatama2016learning} use built-in transition-based parsing module to generate semantic constituency parsing tree for downstream NLP tasks by using RL. 
\citeauthor{yu2017learning}~\shortcite{yu2017learning} propose a RL-based skim reading method, which is implemented on recurrent models, to skim the insignificant time slots to achieve higher time efficiency. 
\citeauthor{choi2017coarse}~\shortcite{choi2017coarse} separately implement a hard attention or a soft attention on a question answering task to generate the document summary.
\citeauthor{shen2017reasonet}~\shortcite{shen2017reasonet} use dynamic episode number determined by RL rather than fixed one to attend memory for efficient machine comprehension.
\citeauthor{hu2017reinforced}~\shortcite{hu2017reinforced} employ policy gradient method to optimize the model for non-differentiable objectives of machine comprehension, i.e., F1 score of matching the prediction with the ground truth.
\citeauthor{li2017end}~\shortcite{li2017end} propose a service dialog system to sell movie tickets, where the agent in RL is used to select which user's information should be obtained in next round for minimum number of dialog rounds to sell the ticket. 
\citeauthor{zhang2017sentence}~\shortcite{zhang2017sentence} simplify a sentence with objectives of maximum simplicity, relevance and fluency, where all three objectives are all non-differentiable w.r.t the parameters of model.

\section{Conclusions} \label{ext:conclusion}

This study presents a context fusion model, reinforced self-attention (ReSA), which naturally integrates a novel form of highly-parallelizable hard attention based on reinforced sequence sampling (RSS) and soft self-attention mechanism for the mutual benefit of overcoming the intrinsic weaknesses associated with hard and soft attention mechanisms. The hard attention modules could be used to trim a long sequence into a much shorter one and encode rich dependencies information for a soft self-attention mechanism to process. Conversely, the soft self-attention mechanism could be used to provide a stable environment and strong reward signals, which improves the feasibility of training the hard attention modules. Based solely on ReSA and a source2token self-attention mechanism,  we then propose an RNN/CNN-free attention model, reinforced self-attention network (ReSAN), for sentence encoding.  Experiments on two NLP tasks – natural language inference and semantic relatedness – demonstrate that ReSAN deliver a new best test accuracy for the SNLI dataset among all sentence-encoding models and state-of-the-art performance on the SICK dataset. Further, these results are achieved with equal or fewer parameters and in less time.

\section*{Acknowledgments}
This research was funded by the Australian Government through the Australian Research Council (ARC) under grants 1) LP160100630 partnership with Australia Government Department of Health and 2) LP150100671 partnership with Australia Research Alliance for Children and Youth (ARACY) and Global Business College Australia (GBCA).
We also acknowledge the support of NVIDIA Corporation and MakeMagic Australia with the donation of GPU.

\bibliographystyle{named}
{\small \bibliography{ijcai18}}

\newpage
\appendix

\section{Comparison to a Iterative Sampling} \label{ext:comparison}

To verify the RSS that uses parallel discrete sampling is sufficient to trim the long sentence and model the dependencies, we implement the iteration-based sequence sampling method following  \citeauthor{lei2016rationalizing}~\shortcite{lei2016rationalizing} and integrate it with the soft self-attention in the same way as ReSA. 

Given a input sequence, $\bm{x} = [x_1, \dots, x_n]$, iterative sampling aims to learn the following product distribution.
\begin{align} \label{eq:z_seq_prob_apd}
&p(\bm{z}|\bm{x}; \theta_r) = \prod_{i=1}^{n} p(z_i|\bm{x}; z_{1:i-1};\theta_r).
\end{align}
A RNN is used to parameterize the conditional probability function above and the basic RNN rather than LSTM or GRU is employed to reduce the number of parameters. The latent state of the RNN can be referred to as the embedding of both contextual information and history selection results. The recurrence can be formally written as
\begin{align}
&p_i = \sigmoid(w^{T}\sigma(W^{(p)}[h_{i-1};x_{i}] + b^{(p)})+ b), \\
&z_i  \sim p_i, \\
&x'_i = [x_i; z_i], \\
&h_i = \rnn(h_{i-1}, x'_i; \theta_{rnn}),
\end{align}
where $\sim$ denotes the discrete sampling operation and  $\theta_{rnn}$ is the learnable parameters of RNN. Consequently, after this recurrence over the input sequence, a sequence of sampling result, $\bm{z} = [z_1, \dots, z_n]$, is obtained, which shares the same notion with RSS.

We then apply two iterative sampling modules which make selections over the dependent and head tokens, respectively. The output of these two sampling modules is formated as a mask which is then applied to the compatibility function of soft self-attention mechanism. The details of the integration are described in the main paper.

For the comparison of RSS and iterative sampling, we also implement the ReSAN with iterative sampling on SNLI dataset that is one of the largest NLP dataset designed to test the sentence-encoding model. A thorough comparison of them in terms of parameters number, training/inference time, training/test accuracy are show in Table \ref{table:comparison_rss_iter}

\begin{table}[htbp] \small
	\centering
	\setlength{\tabcolsep}{4pt}
	\begin{tabular}{@{}lcc@{}}
		\toprule
		& \textbf{ReSAN w/ RSS} &\textbf{ReSAN w/ Iteration}  \\ \midrule
		Parameter Num (300D)&	\textit{3.1m}&	4.0m\\
		Time/Epoch&	\textit{622s}&	2996s\\ 
		Inference Time&	\textit{5.5s}&	17.1s\\ 
		Train Accuracy&	92.6\%&	92.3\%$^*$\\ 
		Test Accuracy&	86.3\%&  86.2\%$^*$ \\ \bottomrule
	\end{tabular}
	\caption{A thorough comparison of a ReSAN with RSS and Iterative Sampling on SNLI dataset. $^*$The accuracies of these two models should be experimentally equal, but, due to the randomness of neural networks (e.g., initialization, batch SGD), there are some experimental error on the accuracies.}
	\label{table:comparison_rss_iter}
\end{table}

As shown in the table, compared with ReSAN with iterative sampling, the one with RSS requires much fewer parameters, $~5\times$ less training time and $~3\times$ less inference time to achieve the competitive test accuracy. This is consistent with the motivation and target for which we develop the RSS.

\end{document}